\documentclass[journal]{IEEEtran}
\usepackage{cite}

\ifCLASSINFOpdf
  \usepackage[pdftex]{graphicx}
\else
\fi
\usepackage{amsmath}
\usepackage{amsfonts}
\usepackage[boxruled]{algorithm2e}
\usepackage{cuted}
\usepackage{makecell}
\usepackage{xcolor}

\usepackage{algorithmic}
\ifCLASSOPTIONcompsoc
  \usepackage[caption=false,font=normalsize,labelfont=sf,textfont=sf]{subfig}
\else
  \usepackage[caption=false,font=footnotesize]{subfig}
\fi
\begin{document}
%
\title{Learning Memory-Dependent Continuous Control from Demonstrations}
%
%
%

\author{Siqing~Hou,
        Dongqi~Han
        and~Jun~Tani
\thanks{S. Hou, D. Han and J. Tani are with the Cognitive Neurorobotics Research Unit, Okinawa Institute of Science and Technology Graduate University, Okinawa, Japan}
\thanks{Corresponding author: Jun Tani. E-mail: tani1216jp@gmail.com}}

%
%

\markboth{Submitted to: IEEE TRANSACTIONS ON COGNITIVE AND DEVELOPMENTAL SYSTEMS (Under review)}{}
%



\maketitle

\begin{abstract}
Efficient exploration has presented a long-standing challenge in reinforcement learning, especially when rewards are sparse. A developmental system can overcome this difficulty by learning from both demonstrations and self-exploration. However, existing methods are not applicable to most real-world robotic controlling problems because they assume that environments follow Markov decision processes (MDP); thus, they do not extend to partially observable environments where historical observations are necessary for decision making. This paper builds on the idea of replaying demonstrations for memory-dependent continuous control, by proposing a novel algorithm, Recurrent Actor-Critic with Demonstration and Experience Replay (READER). Experiments involving several memory-crucial continuous control tasks reveal significantly reduce interactions with the environment using our method with a reasonably small number of demonstration samples. The algorithm also shows better sample efficiency and learning capabilities than a baseline reinforcement learning algorithm for memory-based control from demonstrations.
\end{abstract}


%
\IEEEpeerreviewmaketitle

\section{Introduction}
%
%
%
%

\IEEEPARstart{R}{einforcement} Learning (RL) using neural networks as parametric functions, a.k.a. deep RL, has been successfully applied to a variety of reward-oriented tasks, ranging from pixel-based video games \cite{mnih2015human, mnih2016asynchronous, wang2016sample, kapturowski2018recurrent} to robotic control applications \cite{wang2016sample, lillicrap2015continuous, haarnoja2018soft}. One main advantage of RL is that it enables agents to learn highly adaptive control skills through self-exploration. However, most successes have been achieved in relatively simple, idealized environments. To advance deep RL in real-world applications, such as those potentially encountered by robots, there are still a number of difficulties that must be overcome.

One major challenge is the trade-off between exploration and exploitation, a long-standing dilemma in RL. Many tasks only provide sparse rewards, i.e., a single reward obtained only after a specific set of actions is completed. This often causes significant difficulties in real-life robotic tasks, as interactions between the agent and the environment cannot be accelerated by software; thus, the agent spends an intolerable amount of time engaging in exploration before discovering a reward to improve its policy. A typical state-of-the-art RL algorithm usually executes millions of interaction steps to train an agent player to play an Atari 2600 game or to teach an insect-like robot to walk \cite{mnih2015human, mnih2016asynchronous, wang2016sample, kapturowski2018recurrent, lillicrap2015continuous, haarnoja2018soft}, which is an extremely impractical for real-world applications. The exploration problem appears to be inherent in autonomous agents, and is almost impossible to resolve perfectly using a pure RL scheme. \textit{Reward shaping} is a method that smoothes sparse rewards to hand-coded dense rewards, indicating the distance to the goal state. It is useful to reduce exploration overhead in tasks with sparse rewards. However, in practice, it requires a considerable amount of engineering and experimentation. Otherwise, the local minima trap or unexpected behavior may be introduced by the shaping function \cite{ng1999policy, vevcerik2017leveraging}.

Humans can learn much by imitating demonstrations by others. However, imitation is also misleading when demonstrator behaviors are not optimal, or when imitators lack the ability to replicate demonstrator behaviors \cite{price2003accelerating}. If we examine professional video game players, we observe that they improve their skills not only by learning from how other players play, but also by exploring more effective strategies by themselves \cite{kim2015stage}. Extra cost can be given to the agent if it follows the demonstrator's trajectory, forcing it to rely more on its own exploration \cite{duminy2019learning}. It is also possible to balance imitation and exploration by using imitation to accelerate RL, either with homogeneous actions, in which the agent can replicate demonstrator behavior \cite{price1999implicit}, or with heterogenous actions, in which the agent lacks the ability to follow the demonstrator's action \cite{price2001imitation}.
It is also practical to use imitation with RL even when the demonstrator's performance cannot be evaluated \cite{ho2016generative}. 

Recent studies have focused on integrating imitation throughout the RL process, when demonstration samples can be used not only to help agents learn a naive strategy for achieving a goal, but also to improve adaptability through further self-exploration in different circumstances. This way, it is possible for agents to acquire adaptive skills with a reasonably small number of demonstration samples, even if they are not optimal. This approach, known as \textit{RL from demonstrations} (RLfD), has recently been employed in video games \cite{hester2018deep} and robot control \cite{vevcerik2017leveraging}, and is significantly more efficient than RL without demonstrations. 

Previous studies of RLfD assumed that crucial information necessary for making a decision is fully observable \cite{vevcerik2017leveraging, hester2018deep}. However, this assumption does not hold in real-world environments where unobservable information always exists. A step toward real-world applications must still be developed by exploring new schemes that do not require such an assumption. Decision-making should not be based solely on current observations, but also on memory, and unobservable information should be inferred from historical observations. In deep RL, although it is theoretically possible to combine all historical and current observations, the cost becomes unacceptable due to hardware capacities and the curse of dimensionality \cite{friedman1997bias}. A representation of fixed size is expected to enable practical policy learning. Recurrent neural networks (RNNs) have proven effective at solving partially observable tasks in RL \cite{utsunomiya2008contextual, shibata2015reinforcement,  heess2015memory, vezhnevets2017feudal, al2017continuous, kapturowski2018recurrent}. Well-trained RNNs are able to encode useful historical information by its hidden states, which can be used to make optimal decisions. Although RNNs appear to be practical solutions to challenges in partially observable environments, some outstanding issues remain, such as representational drift and long-term memory dependencies \cite{kapturowski2018recurrent}, despite the fact that training an RNN is usually harder than training a feed-forward neural network \cite{pascanu2013difficulty}. Evidently, it is not sufficient to simply embed an RNN in a deep RL agent.

In this study, we consider applying RLfD to more challenging, partially observable continuous control tasks by proposing a novel algorithm called Recurrent Actor-Critic with Demonstration and Experience Replay (READER). READER does not require demonstrations that follow an optimal strategy. Instead, imperfect demonstrations can be used, and an agent can conduct self-exploration to learn strategies that are more efficient than the demonstrated strategies. The proposed algorithm was tested in three different types of continuous control tasks, in which learning optimal control requires (1) a large amount of exploration due to sparse rewards; (2) inference of hidden information from a trajectory of observations; and/or (3) long-term memorization of previous experiences. Experimental results show that performance can be significantly improved even using imperfect demonstrations by humans. The algorithm is also tested against Recurrent Deterministic Policy Gradient from Demonstrations (RDPGfD), an extension of Recurrent Deterministic Policy Gradient (RDPG) \cite{heess2015memory} to RLfD. The comparison shows that READER has better sample efficiency in all experiments, and better learning capabilities in some RL tasks.

\section{Background}
\label{sec:background}
\subsection{Reinforcement Learning in Partially Observable Environments}
Solving memory-dependent control tasks can be formulated into a \textit{Partially Observable Markov Decision Process} (POMDP) \cite{aastrom1965optimal} framework. A POMDP can be used to define interactions between an agent and the environment into discrete time steps. In a POMDP, $\mathcal{S}$ defines the set of possible environment states. $\mathcal{A}$ is the set of actions the agent can perform at each time step. An initial state $s_0$ is sampled at the beginning by the distribution $p_0(s_0)$, and the environment reacts to the action following a Markovian dynamic $p(s_{t+1}|s_t, a_t)$. A reward function, $r(s_t, a_t)$, defines the immediate reward given to the agent at each time step ($t$). In contrast to MDP, in a POMDP, the agent cannot observe the exact environmental state, $s_t$. Instead it can receive an observation $x_t$ from the observation set $\mathcal{X}$, conditioned on the underlying state through $p(x_t|s_t)$. The current paper assumes that unobservable information can be inferred from historical observations.

Using RNNs as function approximators to estimate the current true state is a common way to overcome these difficulties and has been utilized effectively in many studies \cite{utsunomiya2008contextual, heess2015memory, vezhnevets2017feudal, kapturowski2018recurrent}. In general, RNNs encode the history of observations, $x_{0:t}$, into a vector representation $h_t$ known as the hidden state. At each time step $t$, the new hidden state is calculated using $h_t = f(h_{t-1}, x_t)$ and the output is calculated using $o_t = g(h_t)$. An RNN can be trained by backpropagation through time (BPTT) \cite{werbos1990backpropagation}. In keeping with \cite{heess2015memory, kapturowski2018recurrent}, we use long-short term memory (LSTM) \cite{hochreiter1997long} architecture as the RNN for encoding historical observations in our algorithm.

    \subsection{Reinforcement Learning from Demonstrations (RLfD) }
        \label{sec:related}
        RLfD has been studied in many different cases. Some algorithms have explored the use of an extra action \cite{taylor2011integrating, wang2017improving} to follow the demonstrator. These methods usually require a knowledge of the demonstrator's policy, and work well only if the action space is discrete. \cite{subramanian2016exploration} introduced an exploration policy to enable both self-exploration and exploration from demonstrations. Behavior cloning has also been used to learn from accurate demonstrations \cite{nair2018overcoming, rajeswaran2017learning}. Another set of algorithms has been introduced to enable an agent to learn from demonstrations by putting the state transition samples ($x_t, x_{t+1}, a_t, r_t$) from the demonstrator into a replay buffer, known as \textit{demonstration replay}, and using them to assist policy evaluation and optimization with off-policy RL. Some widely used off-policy RL algorithms, such as deep deterministic policy gradient (DDPG) \cite{lillicrap2015continuous} and deep Q-networks (DQN) \cite{mnih2013playing}, have been extended with this concept of demonstration replay in DDPGfD \cite{vevcerik2017leveraging} and DQfD \cite{hester2018deep}. These methods use a demonstration replay buffer to learn a state-action value $Q$ function, and to train the policy network (as in DDPGfD) or to derive the policy directly (as in DQfD) from the $Q$ function. 
        
        These RLfD studies were all conducted in the framework of MDPs, whereas RLfD for memory-dependent tasks has not been investigated. In this paper, we introduce this same concept of demonstration replay into an RL algorithm, \textit{Sample Efficient Actor-Critic with Experience Replay} (ACER) \cite{wang2016sample} by proposing several modifications to the original algorithm, and we extend it to the POMDP framework by employing an RNN.  In the following section, we describe the application of policy evaluation and policy optimization using a demonstration buffer.
        
\section{Recurrent Actor-Critic with Demonstration and Experience Replay}
\label{sec:method}

    To solve POMDPs where exploration is crucial, we propose an RL algorithm, READER. We demonstrate it by building upon the existing ACER algorithm. Compared to other RLfD algorithms, such as DDPGfD \cite{vevcerik2017leveraging} and DQfD \cite{hester2018deep}, READER learns a memory-dependent policy that can solve POMDPs. READER also uses policy optimization on demonstration samples, making better use of multi-step returns in demonstrations to improve the policy. In this section, we introduce the components of READER and how READER exploits them to solve exploration-crucial POMDPs.
    \subsection{Recurrent Actor-Critic}
        ACER \cite{wang2016sample} is a state-of-the-art actor-critic algorithm that can learn in both discrete and continuous action cases. We focus on continuous action space in this work. To extend ACER for use in POMDPs, we use RNNs to approximate value and policy functions. As in ACER, we use two networks parameterized by $\theta$ and $\theta_v$ called the policy network and the value network. The policy network does not learn a parameterized distribution, $\pi_\theta (\cdot|x_t)$, conditioned on the current state. Instead, it learns a policy conditioned on a full history of observations from the beginning of the episode, $\pi_\theta (\cdot|X_t)$, where $X_t = x_{0:t}$. We will use the notation $X_t = x_{0:t}$ hereafter, and any function with this notation is expected to be computed recursively using an RNN. RNNs are employed to encode the history of observations $x_{0:t}$ into a vector representation $h_t$. $h_{-1}$ is set to a fixed value, and $h_t$ is recursively computed through the following equation: $h_t = f(h_{t-1}, x_t)$. The policy at time $t$ can be computed as a distribution conditioned $h_t$. Similarly, the value network takes the trajectory of observations, $X_t$, as input and then outputs $V_{\theta_v}(X_t)$ and $Q_{\theta_v}(X_t, a_t)$.
        
        To better estimate value functions, we propose an architecture for the value network: the \textit{stochastic recurrent dueling network} (SRDN), which is a recurrent version of the stochastic dueling network \cite{wang2016sample}. Instead of estimating $Q^\pi(X_t, a_t)$ directly, in SRDN, we calculate $A_{\theta_v} (X_t, a_t)$ as an estimate of the \textit{advantage} $A^\pi(X_t, a_t) = Q^\pi(X_t, a_t) - V^\pi(X_t)$, and correct the bias by random sampling $n$ from the policy $\pi_\theta$, leading to the following stochastic estimate of $Q^\pi$: 
        \begin{equation}
        \begin{split}
            \tilde{Q}_{\theta_v}(X_t, a_t)
              = & V_{\theta_v}(X_t) + A_{\theta_v} (X_t, a_t) - \frac{1}{n} 
              \sum_{i=1}^{n}A_{\theta_v} (X_t, u_i), \\
              & \mathrm{where} \: \: u_i \sim \pi_\theta(\cdot| X_t).
        \end{split}
        \end{equation}
        $A_{\theta_v}$ shares the same recurrent layer with $V_{\theta_v}$ in SRDN. This equation will maintain a consistency between $\tilde{Q}_{\theta_v}$ and $V_{\theta_v}$ such that
        $
            \mathbb{E}_{a, u_{1:n} \sim \pi(\cdot, X_t)} \left(
                    \tilde{Q}_{\theta_v}(X_t, a)
                \right) = V_{\theta_v}(X_t).
        $
        
    \subsection{Prioritized Episodic Demonstration and Experience Replay}
        Experience replay has been widely used in reinforcement learning to improve sample efficiency \cite{mnih2013playing, wang2016sample, lillicrap2015continuous} by storing past interactions with the environment into a replay buffer. Past interactions are randomly sampled from the replay buffer and used as training samples, breaking the temporal correlation among samples. Prioritized experience replay (PER) \cite{schaul2015prioritized} is an improvement on experience replay that samples potentially useful interactions with high priority. 
        
        In memory-dependent control, a continuous trajectory of time steps should be sampled as RNN input rather than as independent time steps. \cite{kapturowski2018recurrent, hausknecht2015deep} discussed several ways of sampling a trajectory, including sampling the whole episode. The tasks used in our experiments are relatively short (no more than 200 time steps in length), so we sample and replay the whole episode trajectory. High sample variance is caused by correlation within samples in the same episode \cite{kapturowski2018recurrent}. To prevent this, we sample several times at each learning step and form a batch from the sampled episodes to introduce uncorrelated samples from other episodes. 
        
        We follow \cite{kapturowski2018recurrent} in defining the priority of the $i$th episode, $p_i$, through a mixture of max and mean absolute n-step TD-errors, $\delta_i$, over the episode: 
        \begin{equation}
            p_i = \eta \max_i \delta_i + (1 - \eta) \bar{\delta} .
        \end{equation}
        TD-errors are also recorded in the replay buffer. However, TD-errors change when the policy is updated, and we cannot afford the cost of recomputing TD-errors for all episodes after each learning step. Instead, TD-errors of an episode are only updated after being replayed. The probability of sampling the $i$th episode is then calculated using the following equation: $P(i) = \frac{p_i^\alpha}{\sum_i p_i^\alpha}$. To correct bias introduced by non-uniform sampling, updates to the neural networks are scaled by the following importance sampling weight: 
            $\frac{L(i)}{N} \cdot \frac{1}{P(i)}$.
        $L(i)$ is the length of the $i$th episode, and $N$ is the sum of the lengths of all episodes stored in the replay buffer. To keep a uniform update on each environmental step in the replay buffer, the importance sampling weight is also scaled down by the length of episode $L(i)$, and the resulting weight is used to scale the update of both the policy and value networks:
        \begin{equation}
            \label{equ:weight}
            w_i = \frac{1}{N} \cdot \frac{1}{P(i)} .
        \end{equation}
        
        As in DDPGfD, we store demonstration episodes in a separate prioritized replay buffer called the \textit{demonstration buffer}. The replay buffer containing past experiences is then called the \textit{experience buffer}. In each learning step, sampling is performed by putting the demonstration and experience buffers together in a combined buffer, using the same definition of priority and update rule of TD-errors mentioned above. Probabilities and important sampling weights are calculated with respect to the combined buffer.

    \subsection{Policy Evaluation}
        Policy evaluation is a necessary part of ACER. We extend it using the RNNs mentioned above. The goal is to learn an optimal policy $\pi^{*}(\cdot|X_t)$, which is a distribution of actions conditioned on the history of observations. During the process of improving the current policy $\pi$, the state-action value function $Q^\pi(X_t,a)$ and the state value function $V^\pi(X_t)$ are learned simultaneously. Learning these two value functions is termed policy evaluation. 
        
        Retrace($\lambda$) \cite{munos2016safe} is the policy evaluation method used in ACER. In experience replay, the policy used to generate previous episodes, also known as the \textit{behavior policy} $\mu$, is not the same as the current policy ($\pi$). A set of trajectories sampled from $\mu$ is given to estimate $Q^\pi(X_t,a)$ and $V^\pi(X_t)$. Consider the general operator of return-based policy evaluation used for updating $Q$: 
        \begin{equation}
        \begin{split}
             & \mathcal{R}Q(X_{t_0},a) = Q(X_{t_0},a)+ \\ & \mathbb{E}_\mu 
                            \left[
                                \sum_{t \geq t_0} \gamma^{t-t_0} \left( 
                                    \prod_{s=t_0+1}^t c_s
                                \right)
                                (r_t + \gamma \mathbb{E}_\pi Q(X_{t+1}, \cdot) - Q(X_t, a_t))
                            \right].
        \end{split}
        \end{equation}
        The coefficient $c_s$ is called \textit{traces} of the operator. $\prod_{s=t_0+1}^t c_s$ is defined as $1$ when $t=t_0$. Different $c_s$ choices lead to different policy evaluation algorithms \cite{munos2016safe, harutyunyan2016q, precup2000eligibility, precup2001off}. Retrace($\lambda$) is a policy evaluation algorithm that can be applied to arbitrary behavior policy $\mu$, when $c_s=\lambda \min\left(1, \frac{\pi(a_s|X_s)}{\mu(a_s|X_s)}\right)$ for $0 \leq \lambda \leq 1$. Retrace($\lambda$) has several advantages. It can utilize full returns in near on-policy cases, and does not suffer from high variance in off-policy cases. We can derive a bootstrap equation of Retrace($\lambda$) from one trajectory sampled from $\mu$ to estimate $Q^\pi(X_t, a_t)$: 
        \begin{equation}
        \begin{split}
            & Q^{ret}(X_t, a_t) = r_t + \\ & \gamma \bar{\rho}_{t+1} \left[ 
                Q^{ret}(X_{t+1}, a_{t+1}) - \tilde{Q}_{\theta_v}(X_{t+1}, a_{t+1})
            \right] + \gamma V_{\theta_v}(X_{t+1}).
        \end{split}
        \end{equation}
        
        Retrace($\lambda$) assumes that the behavior policy is known to the algorithm. However, this does not hold in demonstration samples. Demonstration samples, generated by a human demonstrator, do not necessarily follow a parameterized policy. Hence, we cannot use Retrace($\lambda$) to evaluate the policy $\pi$ from demonstrations because $\mu(a|X)$ is intractable. To solve this issue, a one-step estimator $Q^{0}$ \cite{precup2000eligibility} is used to evaluate the policy $\pi$ from demonstrations, by setting the trace $c_s=0$:
        \begin{equation}
            \label{equ:policy_op}
            Q^{0}(X_t, a_t) = r_t + \gamma V_{\theta_v}(X_{t+1}).
        \end{equation}
        Although this does not utilize the full return from the demonstration samples, it does guarantee the convergence of policy evaluation. For experience samples generated by a past policy $\mu$ of the agent, we record $\mu(a_t|X_t)$ and use Retrace($\lambda$) to evaluate the current policy $\pi$. 
        
        $V^\pi(s)$ is also trained with the target 
        \begin{equation}
        \begin{split}
            & V^{target}(X_t) = \\ & \min\left(1, \frac{\pi(a_t|X_t)}{\mu(a_t|X_t)}\right) 
                    \left( Q^{ret} (X_t, a_t) - \tilde{Q}_{\theta_v}(X_t, a_t) \right) + V_{\theta_v}(X_t).
        \end{split}
        \end{equation}
        This is only trained on experience samples, for the same reason that $\mu(a_t|X_t)$ is intractable in demonstration samples.
        
    \subsection{Policy Optimization}
        We follow the use of policy gradient with truncated importance sampling weights in ACER to update the policy network, but without trust region policy optimization \cite{schulman2015trust} and bias correction \cite{wang2016sample} for simplicity. The gradient of network parameter $\theta$ is:
        \begin{equation}
            \hat{g}_t(\theta) = \bar{\rho}_t 
                \nabla_\theta \log \pi_\theta (a_t|X_t) [Q^{target}(X_t, a_t) - V_{\theta_v}(X_t)],
        \end{equation}
        where $\bar{\rho}_t = \min(c, \rho_t)$, and $c$ is a hyperparameter used to truncate the update to the policy network. In experience buffer, $\rho_t = \frac{\pi(a_t|X_t)}{\mu(a_t|X_t)}$, while in demonstration buffer, $\rho_t = \pi(a_t|X_t)$. 
        
        We use $Q^{ret}$ as $Q^{target}$ when replaying the experience buffer, as in ACER, and $Q^{0}$ in Equation (\ref{equ:policy_op}) as $Q^{target}$ when replaying demonstration buffer in order to make use of full returns of the episodes.

    \subsection{Pre-trained Policy Network}
    As an optional step, we use behavior cloning\cite{nair2018overcoming} to pre-train the policy network. Instead of being randomly initialized, the policy network is trained from demonstrations before interacting with the environment, by maximizing the likelihood of following the demonstrations. In our case, the policy network outputs the mean of a Gaussian with fixed variance; thus, it is equivalent to minimizing the squared error between network outputs and recorded actions.
    
    Since environments used in our experiments are randomly initialized, we need enough samples in the demonstrations to generalize different initializations in order to benefit from pre-training. A pre-trained policy network provides a better start for reinforcement learning than random initialization and it improves learning efficiency. 
        
    \subsection{The READER Algorithm}
        Algorithm \ref{algo} summarizes these ideas; thus, the agent is able to learn a policy in a model-free setting from both demonstrations and from its own past experiences. The algorithm also utilizes a separate target network for the value network, which has been used to stabilize the value estimations in several RL algorithms \cite{mnih2013playing, mnih2016asynchronous, lillicrap2015continuous}. Before RL starts, the algorithm trains the policy network by behavior cloning for $e_{Pretrain}$ epochs. $e_{Pretrain}$ can be $0$, which disables pre-training. Then the algorithm interacts with the environment for $e_{Start}$ episodes before starting to learn. At each learning step, the policy network and value network are updated, and a soft update of the target network is also done. 
        \begin{algorithm*}
        \SetAlgoLined
             Initialize demonstration buffer with collected demonstrations $\mathcal{D}$\;
             Initialize an empty experience buffer $\mathcal{E}$\;
             Initialize the policy network $\pi_\theta$\;
             Initialize the SRDN $(V_{\theta_{v}}, \tilde{Q}_{\theta_{v}})$\ and the target Network $(V_{\theta'_{v}}, \tilde{Q}_{\theta'_{v}})$ with $\theta'_{v} = \theta_v$\;
             Pretrain the policy network for $e_{Pretrain}$ epochs\;
             \For{$k \in \{ 1 .. maxEpisode \}$}{
                Interact with the environment using the policy network, generate an episode $e$\;
                Append $e$ to $\mathcal{E}$\;
                \If{$k > e_{Start}$}{
                    \For{$z \in \{1..learningSteps\}$}{
                        Sample a batch of $b$ episodes from the joint buffer $\mathcal{E} \cup \mathcal{D}$\;
                        Update $\theta$ by Equation (\ref{equ:policy_op}) wrt. $w_i$ in Equation (\ref{equ:weight})\;
                        Set $g(\theta) = 0$ and $g(\theta_v) = 0$ \;
                        \For{Episode $i$ in the batch}{
                            Compute $w_i$ by Equation (\ref{equ:weight})\;
                        \eIf{Episode $i$ is from $\mathcal{E}$}{
                                $g(\theta_v) \leftarrow  g(\theta_v) + w_i \nabla_{\theta_v} \sum_{0 \leq t < L(i)} (Q^{ret}(X_t, a_t)-\tilde{Q}_{\theta_v}(X_t, a_t))^2 $ \;
                                $g(\theta_v) \leftarrow  g(\theta_v) + w_i \nabla_{\theta_v}\sum_{0 \leq t < L(i)} (V^{target}(X_t)-V(X_t))^2$ \;
                                $g(\theta) \leftarrow g(\theta) + w_i \sum_{0 \leq t < L(i)} \hat{g}_t(\theta)$\;
                            }{
                                $g(\theta_v) \leftarrow  g(\theta_v) + w_i \nabla_{\theta_v}\sum_{0 \leq t < L(i)}(Q^{TB}(X_t, a_t)-\tilde{Q}_{\theta_v}(X_t, a_t))^2 $ \;
                                $g(\theta) \leftarrow g(\theta) + w_i \sum_{0 \leq t < L(i)} \hat{g}_t(\theta)$\;
                            }
                        }
                        Update $\theta$ and $\theta_v$ with gradient $g(\theta)$  and $g(\theta_v)$, respectively \; 
                        Update the target network: $\theta'_v \leftarrow \beta \theta'_v + (1 - \beta) \theta_v$ \;
                    }
                }
             }
             \caption{READER}
            \label{algo}
        \end{algorithm*}

\section{Experimental Results}
\label{sec:experiments}
    We tested READER in three different simulated control tasks. We show the learning curve of each experimental setting through the average success percentage or the average \textit{return} during the progress of learning. The \textit{return} is defined as the discounted sum of the reward $R=\sum_{t=0}^{t_{max}} \gamma^t r_t$, where $\gamma$ is the discount factor used in READER. To demonstrate the performance of READER against other RLfD methods, we tested RDPGfD, a method based on RDPG that can also use demonstration samples for memory-dependent continuous control. Details of the RDPGfD algorithm can be found in the Appendix. RDPGfD is tested with the maximum number of demonstration samples available in each task, and comparisons are made between READER and RDPGfD with the same number of demonstration samples.
    
   We chose three tasks where we could test READER on different action dimensions, observation dimensions and difficulty in explorations. Mountain Car is a widely used RL task that requires a large amount of exploration. The sequential target-reaching task used in \cite{han2019remaster} divides the task into 3 sequential stages. Each stage requires the agent to complete a sub-goal. This task requires the agent to infer the current stage from historical observations. Finally, to demonstrate the application of our algorithm in robot control, we also performed a simulation experiment on the robot simulator V-REP \cite{vrep2013}. 
   
   While these tasks may be difficult for an RL agent to learn, they are easily grasped by humans after only a few minutes of training. Videos on how human demonstrations were collected can be found in the supplementary materials. However, humans do not react as quickly or as accurately as robots, so we expect robots to learn better policies than humans through RL. Results show that a reasonable number of demonstrations can reduce required interactions with the environment so as to achieve the same level of performance. The number of demonstrations required for each task depends on the difficulty and complexity of the task. 
   
Hyperparameters used in READER are listed in Table~\ref{table:hyperparameters}. 
        \begin{table*}[!t]

          \centering
        
          \begin{tabular}{lll}
        
           \hline
                    \textbf{Hyperparameter} & \textbf{Description} & \textbf{Value}   \\ 
        
            \hline
            
                $\sigma^2$ & Variance of the action distribution & 0.1  \\ \hline
                $H$ & Number of neurons in fully connected layers & 64  \\ \hline
                $HR$ & Number of LSTM units & 64   \\ \hline
                $n$ & Number of sampled actions for SRDN & 16 \\ \hline
                batch size & Number of episodes in one batch & 32 \\ \hline
                optimizer & Optimization algorithm & Adam \\ \hline
                lr\_actor & Learning rate of the policy network & 1e-4 \\ \hline
                lr\_critic & Learning rate of the value network & 1e-3 \\ \hline
                target update ratio & Linear coefficient of updating target network & 0.99 \\ \hline
                $e_{start}$ & Number of episodes before training started & 10\\ \hline
                learning steps & Number of training steps after each episode & \makecell[l]{160 (in POMountainCar) \\ 40 (in SequentialReach) \\ 100  (in SequentialTouch)}\\ \hline
                $\gamma$ & Discount factor  & \makecell[l]{0.995 (in SequentialReach) \\ 0.99 (in others)}    \\ \hline
                $\lambda$ & Discount of trace in Retrace($\lambda$) & 1.0 \\ \hline
                $c$ & Policy update truncation & 1.0 \\ \hline
                $\alpha$ & Scale factor of episode priority & 0.3 \\ \hline
                buffer size & \makecell[l]{Maximum number of episodes\\ in  the experience buffer} & 1000 \\
            \hline
  \end{tabular}

\caption{Values of hyperparameters}
\label{table:hyperparameters}
\end{table*}
    
    \subsection{The Partially Observable Continuous Mountain Car Task}
\begin{figure*}[!t]
\centering
\subfloat[An illustration of the mountain car task.]{\includegraphics[width=1.7in]{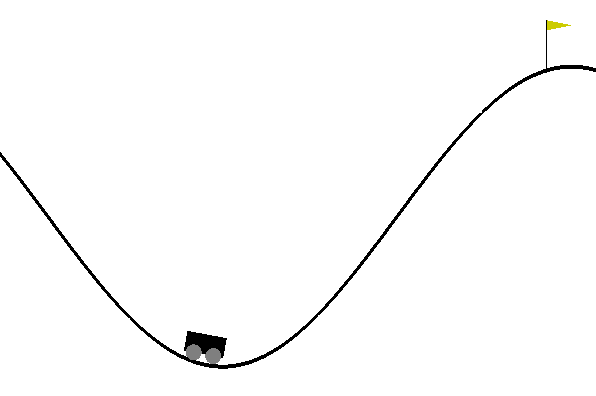}%
\label{fig:MCA}}
\hfil
\subfloat[Learning curves of POMountainCarContinuous with different numbers of demonstrations.]{\includegraphics[width=2.3in]{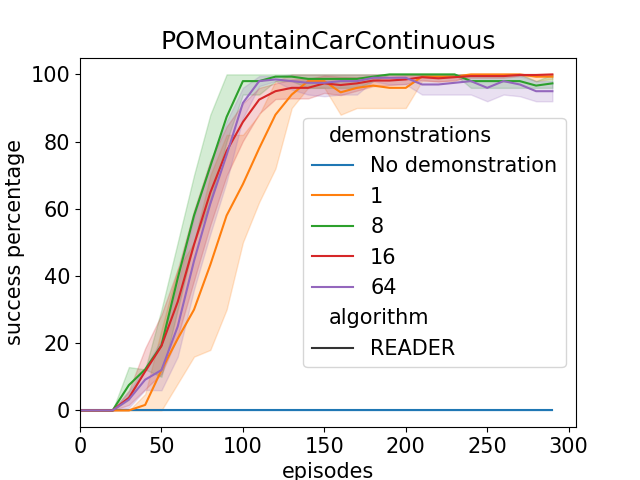}%
\label{fig:MCB}}
\hfil
\subfloat[Learning curves of POMountainCarContinuous when learning from a sub-optimal demonstration, compared to the return of the sub-optimal demonstration.]{\includegraphics[width=2.3in]{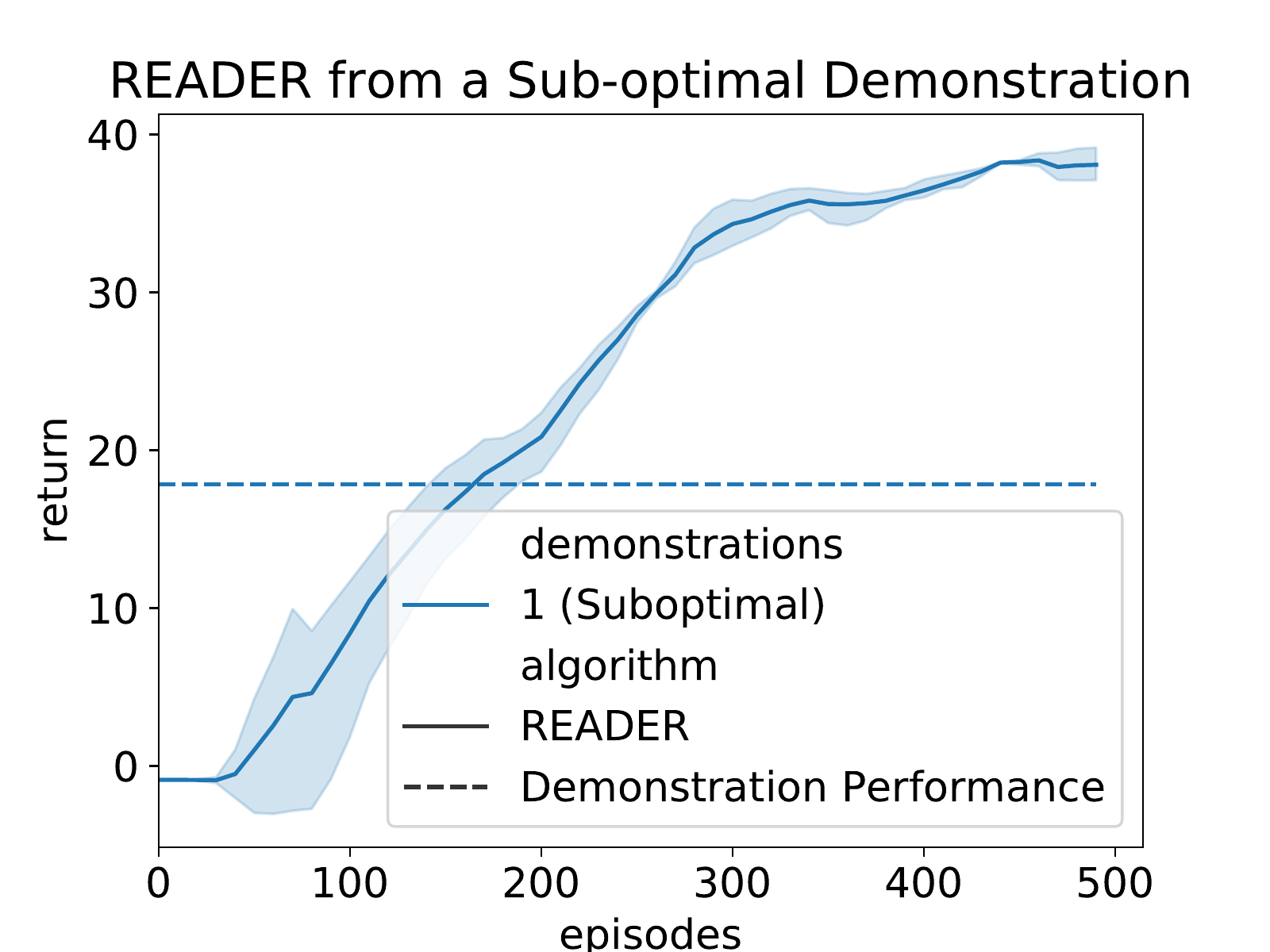}%
\label{fig:MCC}}
\caption{Experiments on POMountainCarContinuous. Error bands in the plots represent the standard errors of the average success percentages at each episode.}
\label{fig:mountaincar}
\end{figure*}

        The mountain car task (Figure~\ref{fig:mountaincar}a) was introduced in \cite{moore1990efficient} as a benchmarking problem of exploration in RL and control. In this task, a car is required to reach a goal site at the top of the mountain on the right. However, it lacks sufficient power to drive straight up the mountain. The only solution is to drive back and forth to build up mechanical energy before ascending to the goal. The agent does not receive any reward until it reaches the goal, making it extremely difficult to find the goal with only autonomous random exploration.
        
        In our setting, the speed of the car is hidden from the agent, and the agent can only observe its own coordinates. We denote this task as \textit{POMountainCarContinuous}. A human is trained to control the car with a joystick, and the environment is displayed as feedback to the human demonstrator. We assume that the agent observes the sensory input in its own view instead of the rendered view that the human demonstrator observes. 
        
        We implemented this task based on the \textit{MountainCarContinous} implementation in OpenAI Gym \cite{brockman2016openai}.
        The time limit is set to 200 time steps. The only change we made to the task is removing speed information from the observations. After this change, the observation is one-dimensional with its horizontal coordinate in the range $[-1.2, 0.6]$. The car starts every episode at a random position from $-0.6$ to $-0.4$ with zero velocity. The action space is $[-1, 1]$. A negative action means pushing the car to the left, while a positive action means pushing the car to the right. The reward is 100 when the car reaches the target of the hill, along with the squared sum of action as a negative reward at every time step.
        
        We collected 4 sets of demonstrations containing 1 episode, 8 episodes, 16 episodes and 64 episodes, respectively. 
        
        Figure \ref{fig:mountaincar}b shows the average return and its standard error over five repeated experiments. The result shows that all 3 sets of demonstrations can lead to near-optimal policies, while learning without demonstrations mostly converges to a local optimum of $0$ return to avoid a negative reward. None of the five agents that learned without demonstrations ever reached the goal, with the consequence that they could never improve their policies. In contrast, even with only 1 episode of demonstration, an agent always quickly improved its policy to reach the goal. 
        
        To test whether the algorithm can learn from sub-optimal demonstrations, we collected another demonstration of 1 episode, during which we asked the demonstrator to circle around the starting position a few times before powering up to the goal. The results (Figure \ref{fig:mountaincar}c) show that the agent learned a better policy than that of the demonstration. The algorithm can work even with sub-optimal demonstrations because it does not directly clone the behavior of the demonstrator. Rather, it extracts reward-related information from demonstration samples to reduce unnecessary exploration. 

    \subsection{The Sequential Target-Reaching Task}
\begin{figure*}[!t]
\centering
\subfloat[An illustration of the SequentialReach task.]{\includegraphics[width=2in]{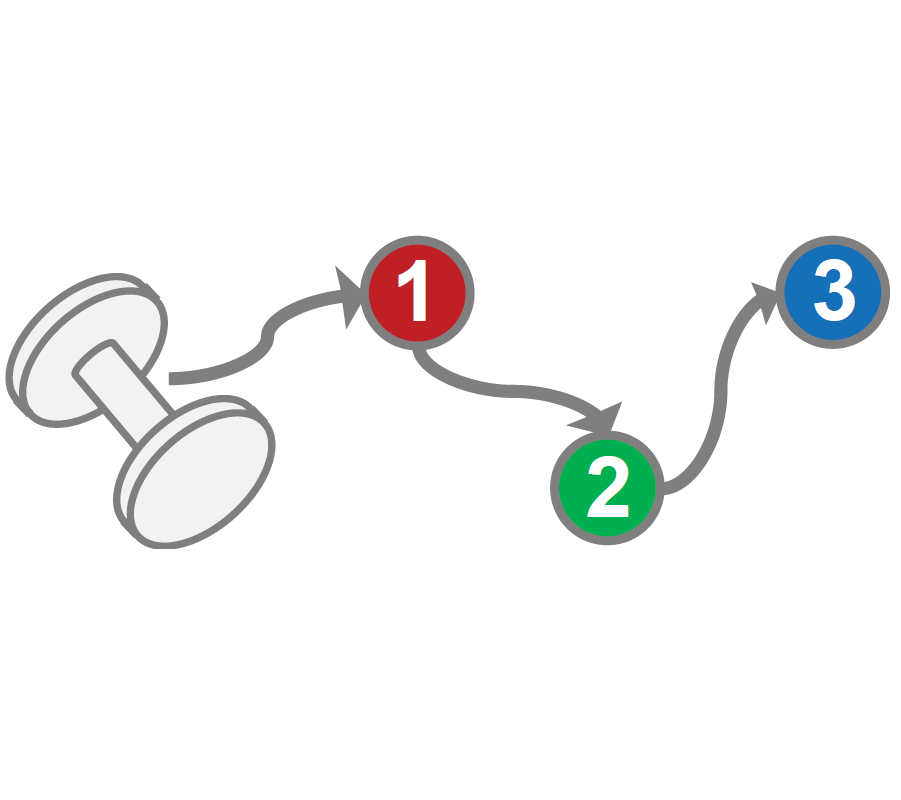}%
\label{fig:T3A}}
\hfil
\subfloat[The rendered view for the demonstrator. The pink triangle with a black tail is the car controlled by the demonstrator. ]{\includegraphics[width=1.7in]{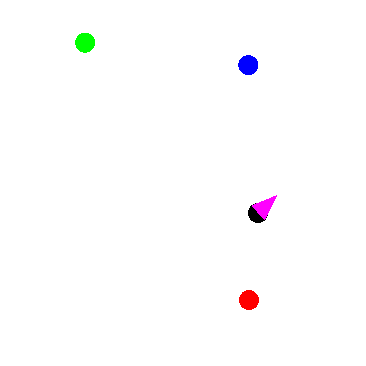}%
\label{fig:T3B}}
\hfil
\subfloat[Learning curves of SequentialReach.]{\includegraphics[width=2.3in]{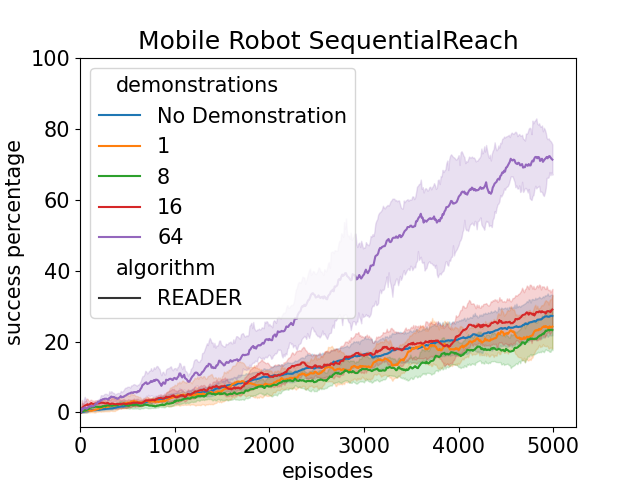}%
\label{fig:T3C}}
\caption{Experiments on SequentialReach. Error bands in the plots represent standard errors of average success percentages at each episode.}
\label{fig:taskt3}
\end{figure*}

        To test our algorithm in memory-crucial environments, we also conducted experiments on the Sequential Target-Reaching task (Figure \ref{fig:taskt3}a) introduced in \cite{han2019remaster}. We denote this task as \textit{SequentialReach}. A two-wheeled robot in a two-dimensional field with a square boundary is required to reach the three randomly sampled target positions in the correct sequence. The robot is rewarded every time it reaches a target in the required sequence (Figure \ref{fig:taskt3}a). The action space is 2-dimensional, representing the rotation speeds of both wheels respectively. The robot can detect distances and angles to the three targets, as well as distances to the field boundary. This task is of interest because there is no clue regarding the current target in the current sensory input, but the agent needs to develop cognitive ability to infer which target to pursue, based on the history of observations.
        
        Our implementation is based on the original example in \cite{han2019remaster}, with modified initial target positions. The robot is placed into a 15 unit $\times$ 15 unit square area, and each of the 3 targets is randomly placed 3 units to 7 units from the center. There is a further restriction that any two targets should be at least $\sqrt{2}$ units away from each other.
        
        As in the previous experiment, a human demonstrator is asked to demonstrate using a joystick. Feedback to the demonstrators is given via a rendered image (Figure \ref{fig:taskt3}b) at each time step. 
        
        The results (Figure \ref{fig:taskt3}c) show that improvements can be achieved in 64 episodes of demonstrations, but even 16 episodes of demonstration do not improve the result significantly. This can be explained by the fact that the configurations of each episodes, such as coordinates of the three targets, are random. The environment cannot be generalized without an adequate number of demonstrations. 

    \subsection{The Robot Sequential Target-Touching Task}
\begin{figure*}[!t]
\centering
\subfloat[A picture of an OP2 and its controlling device.]{\includegraphics[width=1.5in]{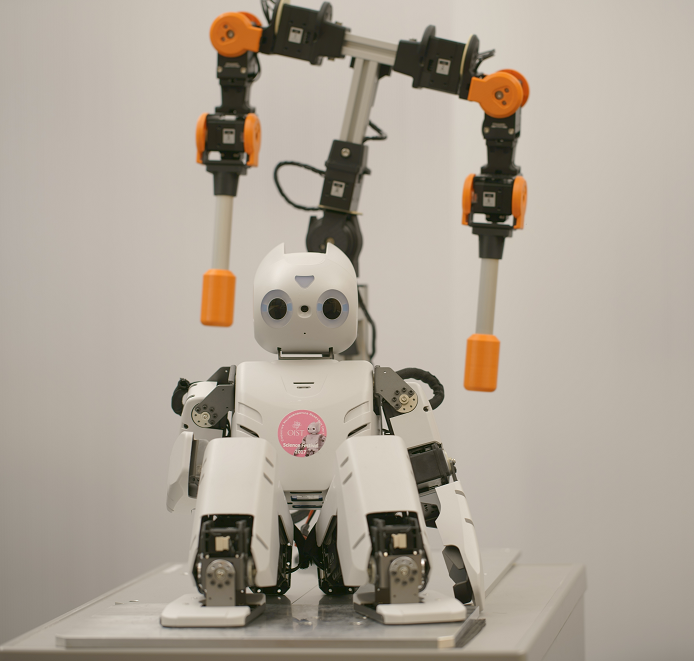}%
\label{fig:OP2A}}
\hfil
\subfloat[A screenshot of the V-REP simulator with the setting of this task. ]{\includegraphics[width=1.43in]{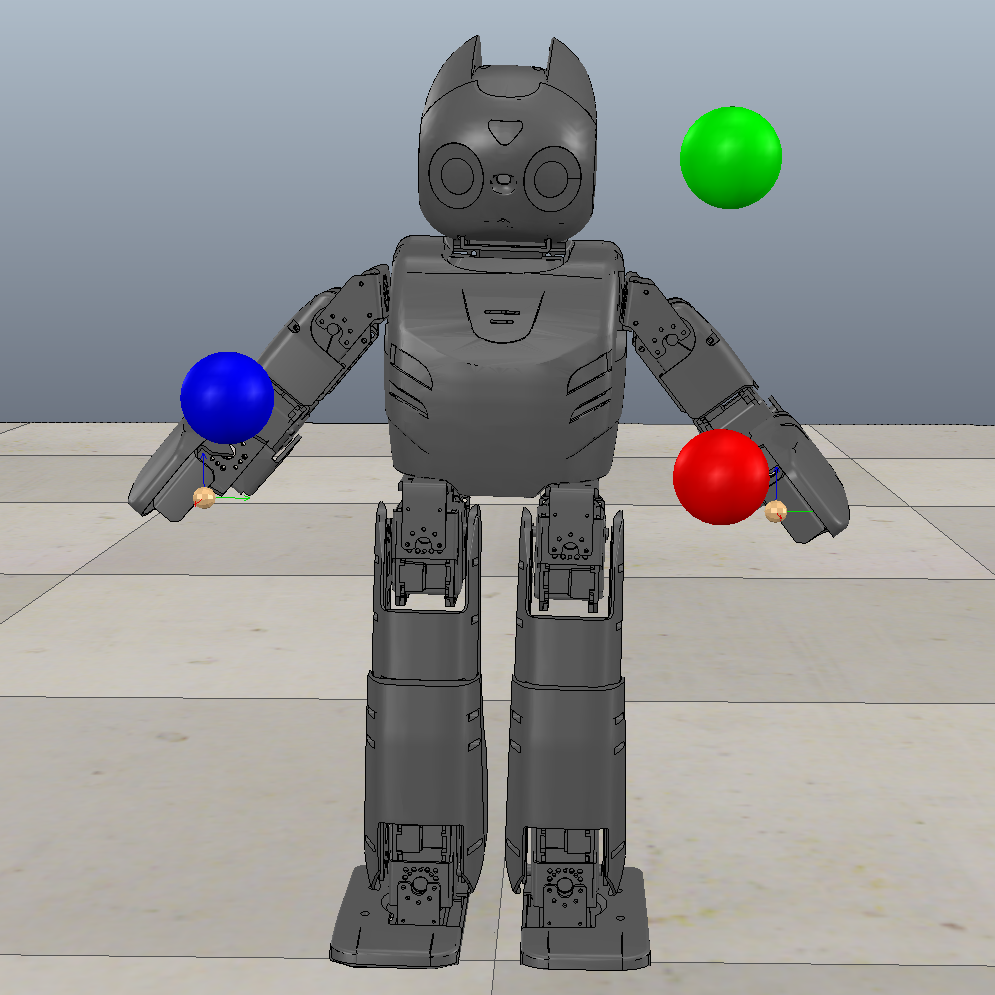}%
\label{fig:OP2B}}
\hfil
\subfloat[Learning curves of SequentialTouch with different numbers of demonstrations. ]{\includegraphics[width=2.0in]{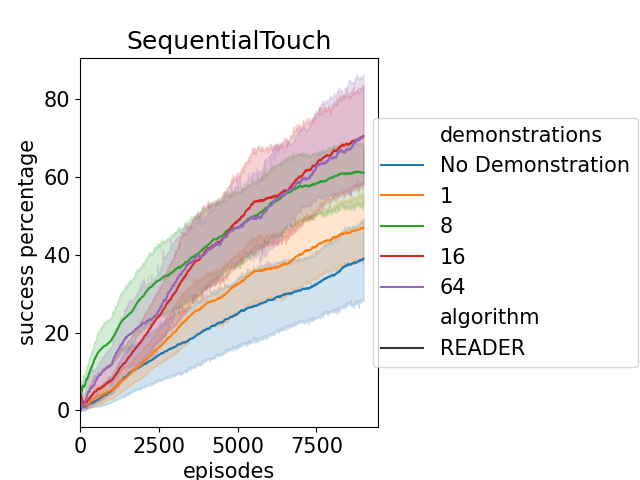}%
\label{fig:OP2C}}
\caption{Simulated humanoid experiment. Error bands in the plots represent standard errors of the average success percentages at each episode.}
\label{fig:OP2}
\end{figure*}

        To demonstrate the efficiency of our algorithm in robot control applications, we introduced a robot simulation experiment on the robot simulator V-REP. We use the simulation model of the humanoid robot OP2 (Figure \ref{fig:OP2}a). We allow the robot to control the 6 motor joints on both arms. However, it is difficult to control a humanoid-like OP2 using joysticks. We used a controlling device (Figure \ref{fig:OP2}a) to control both arms of OP2. We record the joint angles of the 3 joints on the right arm and 3 joints on the left arm, and convert the joint angles to action signals in order to move the simulated robot.
        
        Three spherical targets (one red, one green and one blue) are placed in front of the robot. We introduce some randomness when initializing each episode by sampling the targets at arbitrary positions within a certain range. The robot is required to touch the red ball first, then the green ball, and finally the blue ball with either of both arms. We denote this task as \textit{SequentialTouch}. The robot has sensory input from its 6 joint angles, which it controls, as well as the relative coordinates of each target to the front ends of both arms. The rewards are given when each target is touched in the correct sequence. A reward of 0.5, 2.0 and 1000.0 is given when the robot touches the first target, second target and third target respectively in the correct order. Each time step is 0.2 seconds in real time. The task has a time limit of 50 time steps, which equals 10 seconds in real time.
        
        The results (Figure \ref{fig:OP2}c) show significant improvements using 1, 8, 16 or 64 demonstrations. Results with 8, 16 and 64 demonstrations have similar performance, and are better than result with 1 demonstration.

\begin{figure*}[!t]
\centering
\subfloat[READER compared to RDPGfD in PoMountainCar Continuous]{\includegraphics[width=1.5in]{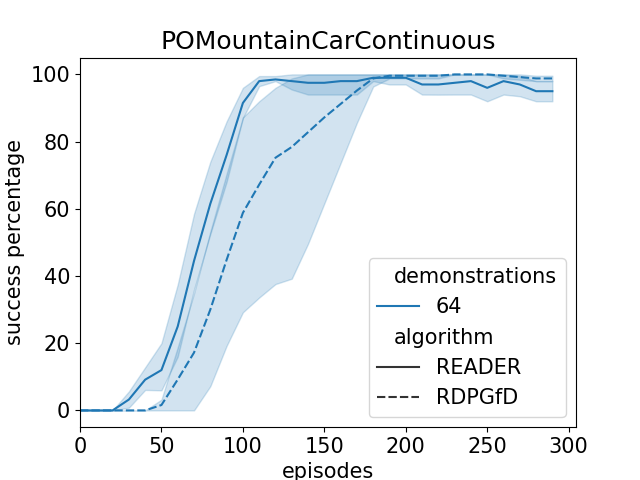}%
\label{fig:fDA}}
\hfil
\subfloat[READER compared to RDPGfD in Sequential Reach]{\includegraphics[width=1.43in]{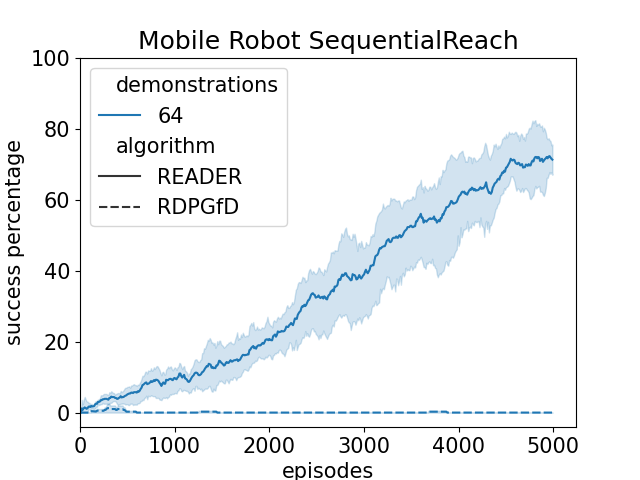}%
\label{fig:fDB}}
\hfil
\subfloat[READER compared to RDPGfD, READER+Pretrain and supervised learning in Sequential Touch]{\includegraphics[width=2.0in]{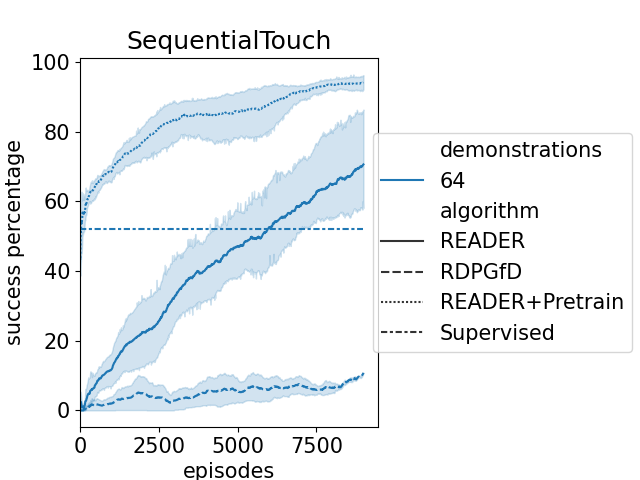}%
\label{fig:fDC}}
\caption{READER compared to RDPGfD and other experiment settings.}
\label{fig:RDPGfD}
\end{figure*} 
    
    \subsection{Comparison to RDPGfD}
    Besides testing READER on the selected environment, we tested RDPGfD in comparison with READER. All tests are done with 64 episodes of demonstrations. In Figure \ref{fig:RDPGfD}, we can see in all three tasks, READER outperforms RDPGfD, either by requiring less explorations or by learning a better policy. 
    
    \subsection{The Effects of Pre-training}
    We also tested a pre-trained policy network using 64 demonstrations. The horizontal line \textbf{Supervised} (Figure \ref{fig:RDPGfD}c) marks the performance of the policy network after pre-training, while the line \textbf{READER+Pre-train} shows the learning curve using the pre-trained policy network. Compared to the pre-trained policy network, RL is necessary after pre-training to achieve better performance on the task. The result with a pre-trained policy network also outperforms the result without a pre-trained policy network in both sample efficiency and convergent success percentage.
    
	\subsection{The Effects of Prioritized Replay}
\begin{figure}[!t]
\includegraphics[width=3.2in]{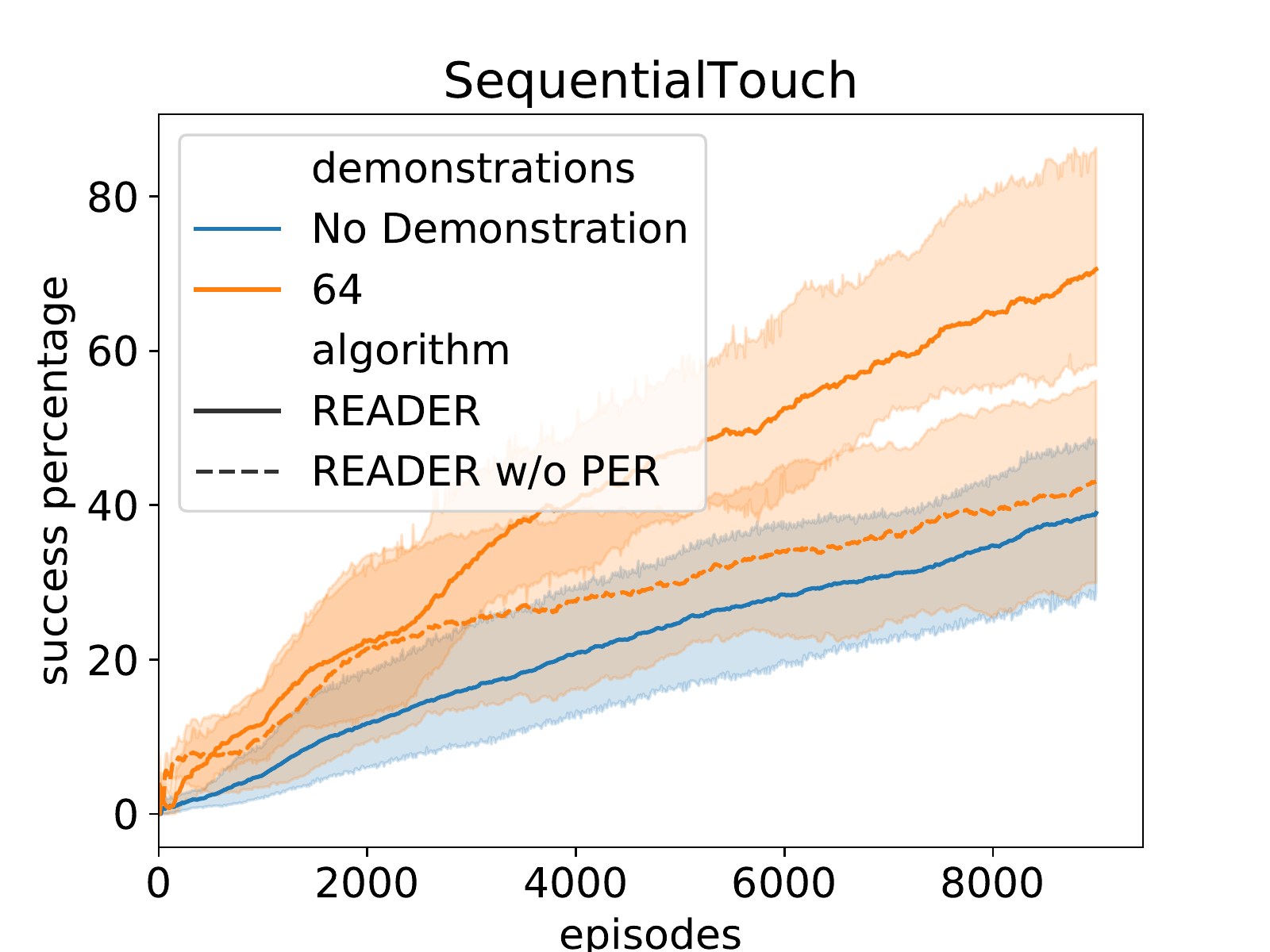}
\caption{The learning curve of READER without prioritized replay compared to READER with prioritized replay and learning with no demonstrations.}
\label{fig:OP2PER}
\end{figure}

\begin{figure}[!t]
\includegraphics[width=3.2in]{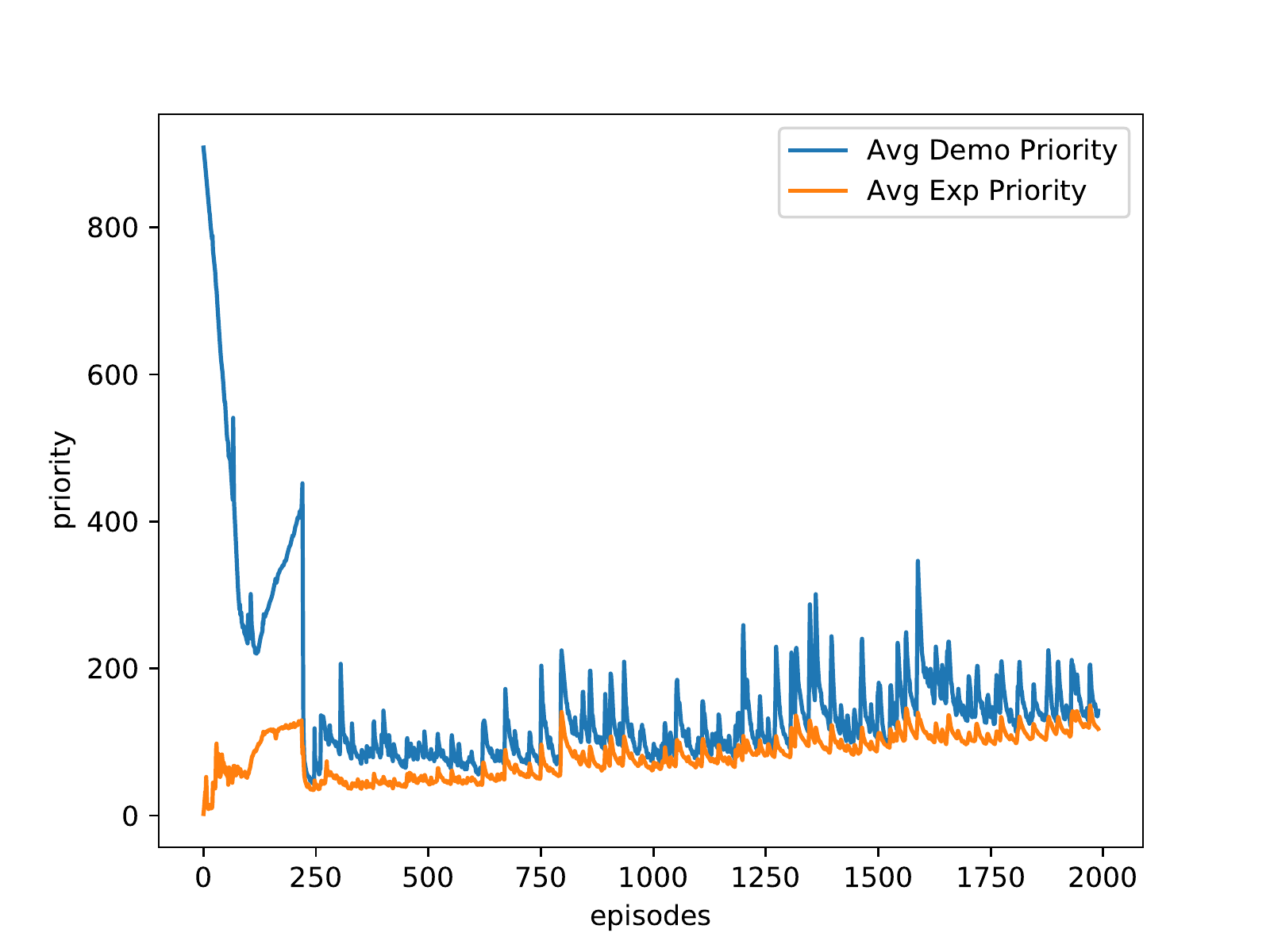}
\caption{Average demonstration priority and experience priority in the first 2000 episodes of SequentialTouch using READER with 64 demonstrations.}
\label{fig:OP2Priority}
\end{figure}

        In order to examine the effects of prioritized episodic demonstration and experience replay (PER), we ran another experiment on SequentialTouch without PER. The results in Figure \ref{fig:OP2PER} show PER contributes significantly to the overall performance. To show how PER affects the attentions of the agents, in Figure \ref{fig:OP2Priority}, we plot the values of average demonstrations priority and average experience priority collected from one run of READER with 64 demonstrations on SequentialTouch. In the beginning 200 episodes of the learning process, average demonstration priority is significantly higher than average experience priority, allowing the agent to learn more from demonstrations. After this period, average demonstration priority and average experience priority converge to close values, while average demonstration priority keeps slightly higher than average experience priority. 
        
\section{Conclusion}
\label{sec:conclusion}

	In this paper, we presented READER, a model-free RL algorithm that uses demonstrations to accelerate learning and to reduce unnecessary exploration. This is crucial for real-world robotic applications because exploration consumes much more time than simulation. READER can be used in memory-dependent POMDPs with continuous action spaces. PER and demonstration replay are used to balance learning with demonstrations and experiences. It also employs RNNs to learn memory-dependent policies. We propose using different update methods of policy evaluation and optimization when using demonstrations and experiences, due to differences between the two sets of trajectories. Our results show that reasonably small numbers of demonstrations significantly improve learning efficiency and reduce the number of episodes required to achieve the same performance. Compared to RDPGfD, another RLfD method for memory-dependent control, READER is more sample efficient and can also solve some tasks that RDPGfD is unable to solve. We also note that the number of demonstrations required to produce a significant performance improvement depends on the difficulty and complexity of the environment. If a task is simple to explore, demonstrations will not improve the sample efficiency significantly since the agent can gather enough successful trials easily through self exploration. Randomness of the environment is a determining factor of required number of demonstrations. If the agent can always follow the same path to finish the task, like in POMountainCarContinuous, then it is expected one episode of demonstration can lead to a significant difference. On the other hand, when the environment is initialized randomly, a larger number of demonstrations should be chosen. A strategy to choose the number of demonstrations is that the demonstrations should cover the major variations of the task, while allowing small differences to be learned further by RL. Furthermore, even when the demonstrations are far from optimal, the agent is still able to outperform the demonstrators by interacting with the environment. Finally, as an optional step, a pre-training policy network from demonstrations can further improve learning efficiency when there are enough demonstration samples to generalize the task.
	
	The main idea of READER is to utilize demonstration samples to achieve more efficient exploration; thus, demonstrations can be sub-optimal. This is especially useful in real-world robot learning tasks because exploration using physical robots can be slow and dangerous. In this sense, our algorithm can be considered as one type of human-robot interaction for developing robot intelligence \cite{aly2016towards, goodrich2008human}. However, it is possible that an agent learns a local-optimal policy if the demonstrator provides a similar one. Efforts to avoid such a predicament will be made in our future work.
	
	Furthermore, future studies may investigate possible performance gains by introducing a model-based approach \cite{igl2018deep, ha2018recurrent} to the current scheme. Such enhanced schemes can allow agents to perform mental simulation while predicting sensory outcomes of their own actions using the learned internal model of the environment. Therefore, agents can explore the environment not only through on-line physical interaction, but also through offline mental simulation. We anticipate that mental simulation along with demonstration will improve sampling efficiency drastically.

\appendix[Recurrent Deterministic Policy Gradient from Demonstrations]
\label{appendix:rdpg}
Recurrent Deterministic Policy Gradient (RDPG) \cite{heess2015memory} is an RL algorithm for memory-based continuous control. We implemented RDPG from Demonstrations (RDPGfD) as a baseline method to compare in our experiments, based on a trivial modification to RDPG that the algorithm samples a batch of episodes from the joint buffer of demonstrations and experiences, just as in the buffer used in READER. RDPGfD uses a policy network $\pi_\theta$ to represent a deterministic policy, as well as a critic network $Q_{\theta_q}$ and its target network $Q_{\theta'_q}$. The target network shares the same structure with the critic network and is updated softly as in READER. All these networks use LSTM, which employs a history of observations as input. To train the critic network, we compute the target of $Q_{\theta_q}(X_t, a_t)$ from $Q_{\theta'_q}(X_{t+1}, a_{t+1})$, the output of the target network at $t+1$: 
        \begin{equation}
            \label{equ:RDPGQ}
            Q^{RDPG}(X_t, a_t) = r_t + \gamma Q_{\theta'_q}(X_{t+1}, a_{t+1}).
        \end{equation}
We compute the policy gradient $\frac{\partial Q_{\theta_q}(X_t, \pi_\theta(X_t))}{\partial \theta}$ to update the deterministic policy network: 
        \begin{equation}
            \label{equ:RDPGpi}
            \frac{\partial Q_{\theta_q}(X_t, \pi_\theta(X_t))}{\partial \theta} = \frac{\partial Q_{\theta_q}(X_t, \pi_\theta(X_t))}{\partial a} \frac{\partial \pi_\theta(X_t)}{\partial \theta}.
        \end{equation}

%





\ifCLASSOPTIONcaptionsoff
  \newpage
\fi



\bibliographystyle{IEEEtran}
\bibliography{bare_jrnl}
\end{document}